\documentclass[preprint,12pt]{elsarticle}

\newcommand{\etal}{et al.~}
\newcommand{\ie}{i.e.~}

\newlength{\smallfigsize}
\makeatletter 
\if@twocolumn
\else
\fi
\makeatother
\usepackage{multirow}
\usepackage{subfigure}
\usepackage{amsfonts}
\usepackage{amsmath}
\usepackage{amssymb}
\usepackage{color}
\usepackage{graphicx}
\graphicspath{{images/}}
\usepackage{url}
\journal{Computer Vision and Image Understanding}

\begin{document}

\begin{frontmatter}

\title{Video Registration in Egocentric Vision under Day and Night Illumination Changes}

\author[unimore]{Stefano Alletto}
\ead{stefano.alletto@unimore.it}
\author[unimore]{Giuseppe Serra\corref{cor}}
\ead{giuseppe.serra@unimore.it}
\author[unimore]{Rita Cucchiara}
\ead{rita.cucchaira@unimore.it}

\address[unimore]{Dipartimento di Ingegneria ``Enzo Ferrari''\\
Universit\`a degli Studi di Modena e Reggio Emilia, Modena MO 41125, Italy}

\cortext[cor]{Corresponding author}

\begin{abstract}

With the spread of wearable devices and head mounted cameras, a wide range of application requiring precise user localization is now possible. In this paper we propose to treat the problem of obtaining the user position with respect to a known environment as a video registration problem. Video registration, \ie the task of aligning an input video sequence to a pre-built 3D model, relies on a matching process of local keypoints extracted on the query sequence to a 3D point cloud. The overall registration performance is strictly tied to the actual quality of this 2D-3D matching, and can degrade if environmental conditions such as steep changes in lighting like the ones between day and night occur.
To effectively register an egocentric video sequence under these conditions, we propose to tackle the source of the problem: the matching process. To overcome the shortcomings of standard matching techniques, we introduce a novel embedding space that allows us to obtain robust matches by jointly taking into account local descriptors, their spatial arrangement and their temporal robustness. The proposal is evaluated using unconstrained egocentric video sequences both in terms of matching quality and resulting registration performance using different 3D models of historical landmarks. The results show that the proposed method can outperform state of the art registration algorithms, in particular when dealing with the challenges of night and day sequences.
\end{abstract}

\begin{keyword}
Video Registration \sep Egocentric Vision \sep Visual matching 
\end{keyword}

\end{frontmatter}


\section{Introduction}

Egocentric vision, thanks to the widespread of cheap and powerful wearable cameras and devices, is increasing its spread among both researchers and consumers. Exploiting the unique first person perspective, many recent works have dealt with the study of self-gestures, social relationships or video summarization \cite{betancourt2014sequential, alletto2015understanding, lee2015predicting}. While this new and unique perspective provides invaluable insights on the viewpoint of the user, challenging situations such as severe changes in the lighting of the environment or high motion blur occur and must be dealt with \cite{betancourt2015evolution}.

A relevant topic that has been recently studied but is yet to be brought to the egocentric field is video registration. That is, the task of precisely localizing an input sequence and, in the case of egocentric videos, the user, with regard to a pre-built 3D model (for example a building of historical interest). A precise estimation of the camera extrinsic parameters in a given timeframe, \ie precise user localization, can be a significant starting point for several egocentric applications such as personalized tours in a city,  assistive services or interactive environments.

The registration of images is a topic that has been widely studied in the past years \cite{LiYunpeng2012, Schindler07, Sattler2012}, on the other hand fewer works have dealt with the registration of video sequences and to the best of our knowledge the employment of egocentric videos has no precedent in literature. In fact, the unique perspective of first person camera views greatly differs from the ones employed in past works under several aspects. For example, fixed camera settings featuring cameras mounted on a van have been exploited, resulting in the acquisition of videos that display very constrained motion patterns and where the rigid setup provides accurate ground truth information about the extrinsic of the cameras used in the testing phase with regard to the ones used to build the Structure from Motion (SfM) model \cite{kroeger2014video, Irschara09fromstructure-from-motion-2009}. On the contrary, egocentric videos often display fast and unpredictable movements and can be acquired under very different conditions from the images or videos used to build the 3D model used in the registration. 


Recent works \cite{Sattler2012,Schindler07} have established a standard pipeline for aligning images or video frames to a pre-build 3D model, which is based on two major steps: feature matching and camera localization. To address the first stage of the pipeline, a widely adopted approach is to extract SIFT feature keypoints and descriptors from a query image and then robustly match them against the descriptors composing the 3D model \cite{Sattler2012}. These correspondences form the 2D-3D matches that will be used to estimate the camera location in terms of rotation and translation matrices, using a Perspective-n-Point (PnP) algorithm often enclosed in a RANSAC loop \cite{Schindler07}.
\begin{figure}[t]
	\centering
		\includegraphics[width=0.80\textwidth]{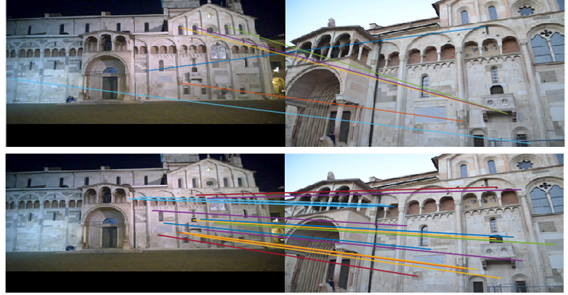}
	\caption{Samples of the matching results. a video frame acquired in a night sequence (left) compared to a model
	image (right). Top: SIFT standard matching technique; bottom: the proposed approach.}
	\label{fig:match_intro}
\end{figure}
The extrinsic parameters estimation, while using algorithms such as RANSAC in order to gain robustness against outliers, strongly depends on the quality of the initial feature matching between the query and the model. In fact, the resulting registration performance decreases if the number and quality of the correspondences found is not sufficient. A major challenge in the video registration from an egocentric perspective derives from the fact that first person videos can span multiples time of the day, and can result in being acquired during the night. Substantial experiments show how matching images acquired during the night against a 3D model built from a collection of images collected in normal, daytime lighting conditions, results in very poor matches, both in terms of quality and number of outliers.

To address the issues deriving from a poor match between query and reference images, recent methods focus on the improvement of the registration results using synthetic views or complex a-posteriori optimizations techniques \cite{kroeger2014video, Irschara09fromstructure-from-motion-2009}. Here, on the contrary, we propose to address the problem at its source and intervene on the matching procedure itself. In particular, we design a novel matching technique that aims at improving the number of scored matches while jointly decreasing the number of outlier. To do so, we propose a novel embedding space that maps local descriptors, its spatial arrangement and temporal robustness of employed keypoints in order to produce a descriptor robust to steep changes in lighting conditions. Our experiments show that this matching technique results in an increase in scored matches in both night and day sequences and in a subsequent improve in registration, without the need of a-posteriori optimization. Figure \ref{fig:match_intro} displays an example of the results obtained by standard SIFT matching on a night-day matching, and compares it with results achieved from our method.

The main contributions of this paper are the proposition of a novel embedding space that takes into account in its design the challenges posed by steep changes in illumination. This embedding space combines local feature descriptors with a representation of their surroundings based on the covariance of densely sampled features. This formulation is further extended to include temporal coherence by tracking local keypoints over a short time to assess their robustness and over-time stability. Finally, we experimentally show that our video registration proposal can cope with the challenges of night sequences with only a small loss in performance and display improved results when compared to current video registration state of the art methods.

\section{Related Work}

Several approaches deal with the task of image registration treating it as an image retrieval problem, matching the query image against a database of images with annotated localization, \ie their rotation and translation matrices aligning them to the desired 3D model \cite{Schindler07, Sattler2012,LiYunpeng2012, zamir2014image, torii2015visual, bourmaud2015wearable}. These approaches tend to be slower due to the high number of comparisons required and can produce a localization that is only accurate at the scale of the single images in the database, but can benefit from established image retrieval methods. Schindler \etal \cite{Schindler07} deal with the task of city scale image localization using a bag-of-words representation of street view images. Similarly, Hay and Efros \cite{Hays2008} compute coarse geo-location information of a query image by matching it to a set of Flickr geo-tagged images. While these approaches can achieve significant performance in scenarios of large-scale localization such as city-scale, their localization is precise at most as the geo-location of the used images and GPS position is often not accurate enough when localizing a camera with regard to a model of a single building.

Aiming at the improvement of localization performance, the use of the 3D structure of the surrounding environment has been recently employed \cite{sattler2011}. In fact, thanks to the recent advancements in Structure from Motion techniques, 3D models can be obtained by a small set of images and can be build on a city-scale with even with consumer computers \cite{wu2011visualsfm}. This results in a shift in paradigm where the descriptors computed on the query image are matched directly to the descriptors of the 3D point-cloud instead of having the intermediate step of matching with the images used to build said point-cloud. To most widely employed descriptors used are the local-invariant SIFT descriptors \cite{lowe2004distinctive}, which are robust to scale variations and to moderate changes in viewpoint. Despite this progresses, the approaches that rely on the matching of interest points succeed only under moderate changes in visual appearance. Image registration in sequences where severe changes in lighting occur due to the acquisition happening during the night still pose a challenge for automated algorithms and especially for their matching phase.

Few approaches have addressed the issues caused by these changes in lighting conditions. Among them, Hauagge \etal \cite{hauagge_bmvc2014_outdoor} design a method based on multiple illumination models, capturing the lighting variations of outdoor environments at different timestamps employing large collections of images of the same outdoor location. On a different note, the work by Torii \etal \cite{Torii2015} proposes a method that deals with the place recognition task using a combination of synthesized virtual views modeling buildings under different lighting conditions and viewpoints and performing dense keypoint matching from the query to the most similar (in terms of viewpoint and lighting) synthetic view available. In contrast to these approaches, we propose a method capable of finding 2D-3D correspondences thanks to a more robust matching phase, that results in a more accurate localization when compared to the use of synthetic models.

The task of registering a video sequence to a 3D model can be performed in two fundamentally different ways. The first, while not being a proper registration technique in the sense that the 3D model is not pre-built but learned online, is Simultaneous Localization and Mapping (SLAM) \cite{ForsterPS14, engel14eccv,concha2015dpptam}. SLAM jointly builds the 3D model of the scene and locates the camera in the environment using the 3D model as reference. Similarly to video registration, this process is often performed through robust keypoint matching; on the other hand a major difference is due to the fact that the camera employed in building the model and in acquiring the query frames is the same, so the internal camera parameters remain constant throughout the problem and both the model and query sequence share the same lighting and environmental conditions. Widely popular in the field of robotics, few SLAM approaches have been extended to the task of video registration.

The problem commonly referred to as video registration employes a pre-build 3D model that can be the result of a collection of heterogeneous images and videos. To effectively register an input video sequence to a model, Zhao \etal \cite{Zhao2004} propose to build a SfM model from the input query and then rigidly aligning this model to the reference 3D point cloud. On the other hand, Lim \etal \cite{LIM-CVPR12} develop a real-time registration technique that uses direct 2D-3D feature matching interleaved with 2D keypoint tracking to increase the robustness of the matching. Irschara \etal \cite{Irschara09fromstructure-from-motion-2009} deal with the problem of registering a video sequence by increasing the number of views and thus of potential keypoint matches by augmenting the model with synthetic views. Recently, the paper by Kroeger \etal\cite{kroeger2014video} demonstrates how direct application of classical registration techniques to the task of registering a video sequences, namely image registration applied to the single frames, results in a noisy localization due to the fact that individual errors can overcome the actual changes in camera location. The authors propose to employ a standard image registration approach on the single frames and then refine it by a global a-posteriori optimization that relies on techniques such as spline smoothing, kernel regression or least squares minimization. The temporal smoothness inherent to the video sequence is exploited thanks to the a-posteriori optimization, allowing the authors to achieve state of the art results.

While current state of the art approaches rely on different kinds of synthesis or a-posteriori refinement in order to improve the registration performance, in this paper we propose to intervene on the feature matching process in order to produce reliable 2D-3D correspondences that can be exploited by PnP algorithms to produce better extrinsic matrices. This is in contrast to obtaining noisy rotation and translation matrices and then optimizing their locations using empirical constraints. In our method the spatio-temporal consistency that is part of a video sequence is exploited by embedding points into a space that produces a representation robust to severe changes in lighting conditions.

\section{The Proposed Approach}
\label{sec:approach}

\begin{figure*}
	\centering
		\includegraphics[width=1\textwidth]{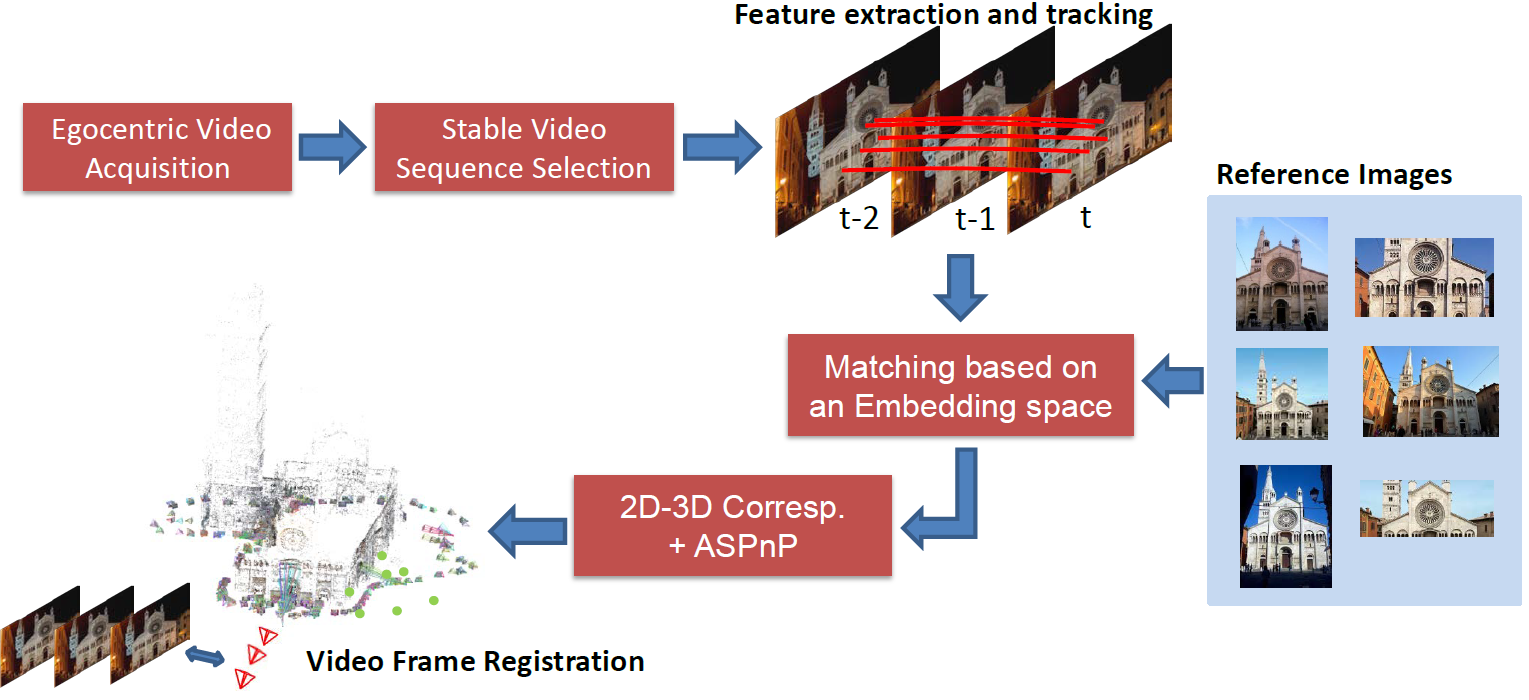}
		\caption{Schematization of the proposed video frame registration approach.}
	\label{fig:method_general}
\end{figure*}

Given an egocentric video sequence, the proposed solution aligns it to a pre-built 3D model generated by performing a SfM method on a set of images.  The SfM pipeline \cite{wu2011visualsfm} first performs image matching to estimate information about the scene structure. In this first step local SIFT features are used, because they have been proved to be able to identify discriminative local elements with good invariant properties to photometric and geometric transformations. After finding a set of geometrically consistent matches between each image pair, the set of camera parameters (position, orientation and focal length) and 3D location for each match is computed by solving a non linear optimization problem that minimizes the reprojection error (the sum of distances between the projections of the 3D Point and its corresponding keypoints).
To build 3D models state of the art techniques exploit a large set of images captured during day time, in which the keypoint extraction and matching process can achieve its best results \cite{kroeger2014video, sattler2011}. Therefore, to build a stable 3D model, we collect images presenting the same characteristics of the ones used in the aforementioned works, that is good lighting conditions.

In egocentric scenarios the first step is to prune the sequence removing unstable frames (fast head movements and physically traveling form a point to another). To predict these events we extract for each frame a visual descriptor based on apparent motion and blurriness. 
Specifically, the visual motion feature is based on optical flow estimated using the Farneback algorithm on a $3 \times 3$ grid. Considering the optical flow $V_x$, $V_y$ (gradients computed for horizontal and vertical components), we compute the motion histogram by concatenating the apparent motion magnitudes $M= \sqrt{V_x^2 + V_y^2}$, with the orientations $\theta= \arctan(V_y/V_x)$, both quantized in eight bins for each frame section, weighting them by their respective magnitude. 
To assess the frame quality, we compute a blur feature using the method presented in \cite{Roffet07}.
Based on these features we train a one-vs-all linear classifier allowing to efficiently remove frames where the presence of motion and blur would otherwise prevent the use of local keypoint descriptors. 


Once that a subset of a video is selected, we retrieve 2D-3D correspondences by matching its frames to the images employed in the model construction and retrieve their correspondences on the 3D point cloud. 
To efficiently retrieve these correspondences we adopt the approach presented in \cite{philbin2007object} to build a ranked list of similar images comparing the image representation of the input query and of the images used in the 3D model building.
After this initial retrieval phase, we propose to refine its results by introducing our matching strategy that involves the first K top ranked images (we experimentally fix K = 25).
Once 2D-3D correspondences are obtained, the absolute camera pose of each frame is determine by solving the perspectiven-point(PnP) problem \cite{Hartley2004, Irschara09fromstructure-from-motion-2009, kukelova13, Yinqiang13}. In our approach we use the ASPnP n-point-pose algorithm \cite{Yinqiang13} enclosed in a RANSAC loop to obtain robust and accurate frame registration, \ie the rotation and translation matrices. Figure \ref{fig:method_general} summarizes the overall pipeline of the proposed method.

\begin{figure*}
	\centering
		\includegraphics[width=0.95\textwidth]{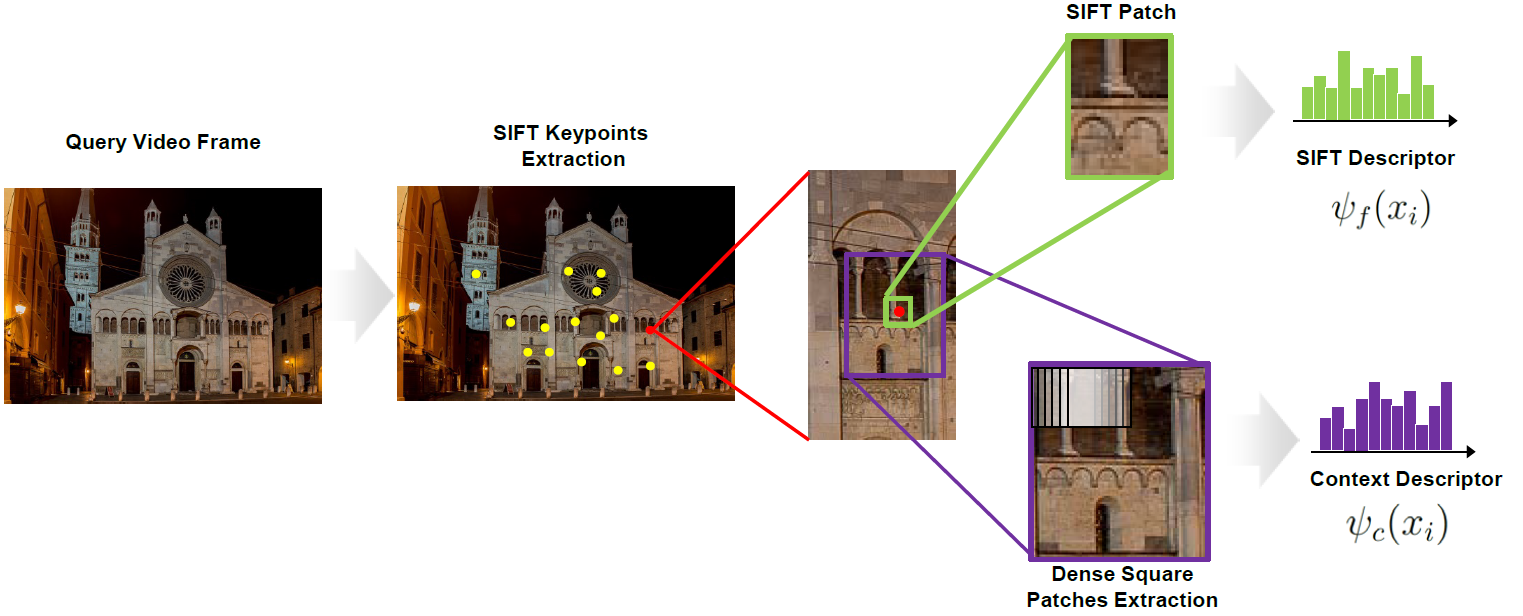}
		\caption{Schematization of the extraction of our descriptors based on SIFT and covariance descriptors.}
	\label{fig:descriptor}
\end{figure*}

Egocentric videos, by their very nature, are recorded in unconstrained scenarios with extreme lighting variations, for example videos can be acquired during day or night time. In this context, standard video frame registration techniques archive poor performance. This is mainly due to the fact that the matching step is not able to select robust features; we hence propose an approach that can identify robust matches even if there are severe changes in illumination conditions between the query sequence and the 3D model.
Figure \ref{fig:descriptor} shows a schematization of the feature extraction.

Based on experimental results, we observed that the use of local feature descriptors only (e.g. SIFT) is not sufficient to obtain accurate matches across large changes in scene appearance due to day/night illumination. In fact, the quality and the number of SIFT matches with significantly different lighting conditions is not enough for effective video registration applications.  Therefore, we propose to represent each keypoint by its local descriptor and its context. 
Formally, an interest point $x_i$ is defined as $x_i=(\psi_g(x_i),\psi_o(x_i),\psi_s(x_i),\psi_f(x_i),\psi_c(x_i) )$ where the symbol $\psi_g(x_i) \in \mathbb{R}^2$ stands for the $2D$ coordinates of $x_i$,  $\psi_o(x_i)$ denotes the orientation information,  $\psi_s(x_i)$ is the scale factor, $\psi_f(x_i) \in \mathbb{R}^D$ corresponds to the local descriptor ($D$ equal to $128$, i.e. the coefficients of the SIFT descriptor), while $\psi_c(x_i)$ is a representation based on descriptors densely sampled around of $x_i$.  

To obtain $\psi_c(x_i)$, we first consider a square region of interest (RoI) surrounding $x_i$ obtained by multiplying a constant value $\epsilon$ to the scale $\psi_s(x_i)$ (we empirically fix $\epsilon = 6$ based on preliminary experiments).
Let $\textbf{R} = \{\textbf{r}_1 \dots \textbf{r}_N\}$ be a set of local  SIFT features densely extracted on this region, we summarize them by computing their covariance matrix descriptor $\mathbf{C}$: 


\begin{equation}
\mathbf{C} = \frac{1}{N-1} \sum_{i=1}^N (\textbf{r}_i - \textbf{m})(\textbf{r}_i - \textbf{m})^T, 
\end{equation}

where $\textbf{m}$ is the mean vector of the set $R$. This covariance representation,  that encodes information about the variance of the features and their correlations, does not need a visual codebook (required for several image descriptors such as Fisher Vector \cite{perronnin10}, VLAD \cite{jegou10}, LLC \cite{wang10} and more), thus removing the dependence from the specific dataset \cite{bepponegold}. It has been shown in \cite{philbin08} that the accuracy of image retrieval systems based on a visual vocabulary drastically drops if the visual words are extracted using a dataset (Oxford dataset\cite{OxfordDataset}) and the test is performed on another similar dataset (Paris \cite{ParisDataset}). In fact, methods based on the generation of a codebook split dense regions of the descriptor space arbitrarily according to the SIFT distribution on the dataset. Therefore the bins do not equally split the unit hypersphere which SIFT covers, resulting in an uneven distribution of points that could not be the right representation for a different scenario. Since our method is designed to cope with egocentric videos which are inherently unconstrained, not being tied to a training dataset is a necessary working condition. Furthermore, covariance descriptors are reported being more robust than directly exploiting features based on gradients because, under illumination variations, variations of gradients inside a region change less than the gradient intensities themselves \cite{bkak2012learning}.

Although the covariance representation is independent of a specific dataset, the distance between two descriptors can not be computed as a Euclidean distance. In fact, these matrices lie in a Riemannian manifold which is not a vector space. Therefore, a mapping function to a Euclidean space is required. We propose to use the projection transformation from the Riemannian manifold to a Euclidean tangent space, called Log-Euclidean metric. In fact, this manifold is a topological space that is locally similar to a Euclidean space, in which each tangent space has an inner product \cite{tuzel_pami}. 

First the covariance matrix is projected from the Riemannian manifold on a Euclidean space tangent through a tangency matrix $\textbf{H}$. Then, the the projected vector is transformed in orthonormal coordinates. More formally the projection of  $\textbf{C}$ on the hyperplane tangent to $\textbf{H}$ is defined as:

\begin{equation}
\hat{\textbf{C}} = \textrm{vec}_\textbf{I}\left( \log\left( \textbf{H}^{-\frac{1}{2}} \textbf{C} \textbf{H}^{-\frac{1}{2}} \right) \right),
\label{eq:r2e}
\end{equation}

where $\log$ is the matrix logarithm operator, while the vector operator $\textrm{vec}_\textbf{I}$ of a symmetric matrix $\textbf{W}$ on the tangent space at identity $\textbf{I}$ is represented as:

\begin{equation}
\textrm{vec}_\textbf{I}(\textbf{W}) = \left[ w_{1,1} \ \sqrt{2}w_{1,2} \ \sqrt{2}w_{1,3} \ldots w_{2,2} \ \sqrt{2}w_{2,3} \ldots w_{d,d} \right].
\end{equation}

By computing the sectional curvature of the Riemmanian manifold it is possible to show that this space is almost flat \cite{tosatoPAMI2013}. This characteristic of the feature space is particularly suitable when dealing with a local embedding function such as the one that we propose use in order to find matches. In addition, it also allows us to chose $\textbf{H}$ equal to the identity matrix, thus avoiding the parameter optimization for a specific scenario.
As the projected covariance to Euclidean space is a symmetric matrix of $D \times D$ values (e.g. $D=128$, the SIFT dimensionality), the context descriptor $\psi_c(x_i)$ is a $(d^2+d)/2$-dimensional feature vector. 

Based on these features we present a new robust feature matching technique that maps both local and context descriptors in a common embedding space.
Let ${F}=\{(\psi_f(x_i), \psi_c(x_i)\}_{i=1}^p$, $M=\{(\psi_f(y_i), \psi_c(y_i)\}_{i=1}^q$ be respectively the list of interest points features (local and context descriptors) taken from a query video frame and an image registered to pre-built 3D model. To project the feature points ($F$ and $M$) to the common embedded points $Z$, we propose to minimize the following objective function: 

\begin{equation}
Z = \textit{arg} \operatornamewithlimits{\textit{min}}_{Z} \sum_{i,j}  \left\| z_i - z_j \right\|^2 P_{ij}^{FM} R_{ij}^{FM} 
\label{eq:singleframe}
\end{equation}

where $i = 1, \dots, p$, $j = 1, \dots, q$, the weight matrix $P^{FM}$ (where $P_{i,j}^{FM} = K((\psi_f(x_i),(\psi_f(y_j))$) encodes the similarity between local descriptors, while $R^{FM}$ (where $R_{i,j}^{FM} = K((\psi_c(x_i),(\psi_c(y_j))$) represent the similarity between context descriptors. In both cases Gaussian kernel $K$ is adopt, for example 
\begin{equation}
K(\psi_f(x_i),(\psi_f(y_j)) = e^{-\left\| \psi_f(x_i) - \psi_f(y_j) \right\|^2 /{\sigma^2}}
\end{equation}
where $\sigma$ is empirically fixed. A unit L2-norm constraint on $Z$ is applied to avoid trivial solutions.

The minimization of this embedding function maps the points $z_i$ and $z_j$ close in the embedding space if their feature similarity kernels $P_{i,j}^{FM}$ and $R_{i,j}^{FM}$ both present high values.
Notice that a weighted Kronecker product between these two matrices could also be adopted to equalize their similarities. However, the accuracy of the matches with learnt weights tends to be similar to the one obtained using uniform weights chosen through cross-validation \cite{varma2009more}. In addition, it would lead to a dataset-dependent tuning, again in contrast with our purposes; therefore we use uniform weights.  
The resulting embedding function is very suitable, because it can be efficiently solved as an Eigen-Value problem and can be easily extended including spatial and temporal constrains (see next Section). 

The Eq. \ref{eq:singleframe}, following the approach proposed in \cite{Belkin03}, can be reduced to:

\begin{equation}
\begin{aligned}
\hat{Z} = &\textit{arg}\operatornamewithlimits{\textit{min}} \quad Z^TLZ \\
& \text{subject to  }  Z^TDZ=I \quad \text{and} \quad  Z^TD\textbf{1}=I
\end{aligned}
\end{equation}

where $L$ is the Laplacian of the matrix $PR^{FM}= P^{FM} \otimes R^{FM}$; $L=D-PR^{FM}$, where $D$ is the diagonal matrix defined as $D_{ii} = \sum_j PR_{ij}^{FM}$. The constraint $Z^TDZ=I$ has been introduced to remove the arbitrary scaling. Minimizing this function can be done as an eigenvector problem: $L_z = \lambda Dz$, in which the optimal solution can be obtained by the bottom $d$ nonzero eigenvectors.  

\section{Spatial and temporal constraints}
\label{sec:temp}

The aforementioned embedding function (Eq. \ref{eq:singleframe}) can be extended to include time consistency by tracking local keypoints and add spatial constraints by considering their spatio-temporal arrangement. 

Let ${F^{T-k}}=\{(\psi_g^{T-k}(x_i), \psi_f^{T-k}(x_i), \psi_c^{T-k}(x_i)\}_i, \; k=0 \dots K$, and $M=\{(\psi_f(y_i), \psi_c(y_i)\}_{i}$ be the list of the point descriptors of a short-time sequence of $K+1$ frames and $M$ is an image linked to the 3D model as previously described. To include temporal consistency, interest points are tracked through the sequence. Due to the simple nature of short term keypoint tracking, we use the KLT approach \cite{Tomasi91detectionand} which is fast and robust. Interest points that cannot be effectively tracked across the entire sequence are discarded. 

The objective function that we propose to minimize to build the embedding space including the aforementioned constrains is the following:

\begin{equation}
\begin{split}
Z & = \textit{arg} \operatornamewithlimits{\textit{min}}_{Z} \sum_{i,j}  \left\| z_i^T - z_j^M \right\|^2 P_{ij}^{F^TM} R_{ij}^{F^TM} + \\ & + \sum_{i,j}  \left\| z_i^T - z_j^T \right\|^2 S_{i,j}^{T} G_{i,j}^{T} 
\end{split}
\label{eq_temporal_spatial}
\end{equation}

where the matrix $S^T$ and $G^T$ encodes the spatial and temporal similarity respectively: 

\begin{equation}
S_{i,j}^T = e^{-\left\| \psi_g^T(x_i) - \psi_g^T(x_j) \right\|^2 /{\sigma^2} }
\end{equation} 

\begin{equation}
G_{i,j}^T = e^{- \frac{  \sum_{p=1}^{K} \left( \left\| \psi_g^{T-p}(x_i) - \psi_g^{T-p}(x_j) \right\| - \left\| \psi_g^{T}(x_i) - \psi_g^{T}(x_j) \right\| \right)^2}{ {\sigma^2}} }
\label{eq:tempo}
\end{equation}

where  $\psi_g(x_i)$ represents for the $2D$ coordinates of $x_i$. In other words, $S^T$ constrains the embedding space to take into account the spatial arrangement of the interest points of the frame T, while $G^T$ encodes how the spatial arrangement remained constant in the sequence. The objective function in Eq. \ref{eq_temporal_spatial} can be formulated, following the solution presented in \cite{Torki2010}, as:   

\begin{equation}
Z = \textit{arg} \operatornamewithlimits{\textit{min}}_{Z} \sum_{O=\{T,M\}} \sum_{i,j}  \left\| z_i^T - z_j^O \right\|^2 U_{i,j}^{FM} 
\label{eq:reduced}
\end{equation}

where $U^{FM}$ is defined as: 

\begin{equation}
U^{FM} =
  \begin{cases}
    P_{ij}^{F^TM} R_{ij}^{F^TM}   & \quad for \quad M\\
    S_{i,j}^{T} G_{i,j}^{T} 	  & \quad for \quad T\\
  \end{cases}
\end{equation}

Similarly to Eq. \ref{eq:singleframe} this objective function, encoding spatial and temporal constrains (Eq. \ref{eq:reduced}), can be solved as an Eigen-Value problem. The resulting embedding space guarantees that the Euclidean distances between the embedded points take into account both spatial and temporal constraints. Therefore, to select robust matches we can treat it as a bipartite graph matching problem.  That is, when matching two images the two disjoint sets of keypoints, whose description lies in the embedding space, can be arranged into two disjoint sets and the matching problem can be modeled as treating the keypoints as nodes of a bipartite graph. In particular we exploit the Hungarian Algorithm \cite{papadimitriou1982} to find a possible correspondence of each interest point in the frame $T$.This allows the method to find an optimal solution, avoiding the greedy association problem typical of nearest-neighbor matching, which can fail in complex scenarios such as the one tackled by our method. 
Since the Hungarian algorithm always produces a match, to reduce outlier matches we compare the similarity of the candidate match to the one of the second-closest neighbour. If the closest neighbor is significantly closer than the second feature point (following the implementation of \cite{lowe2004distinctive}) the match is considered correct. 

\begin{figure*}[t]
	\centering
		\includegraphics[width=0.95\textwidth]{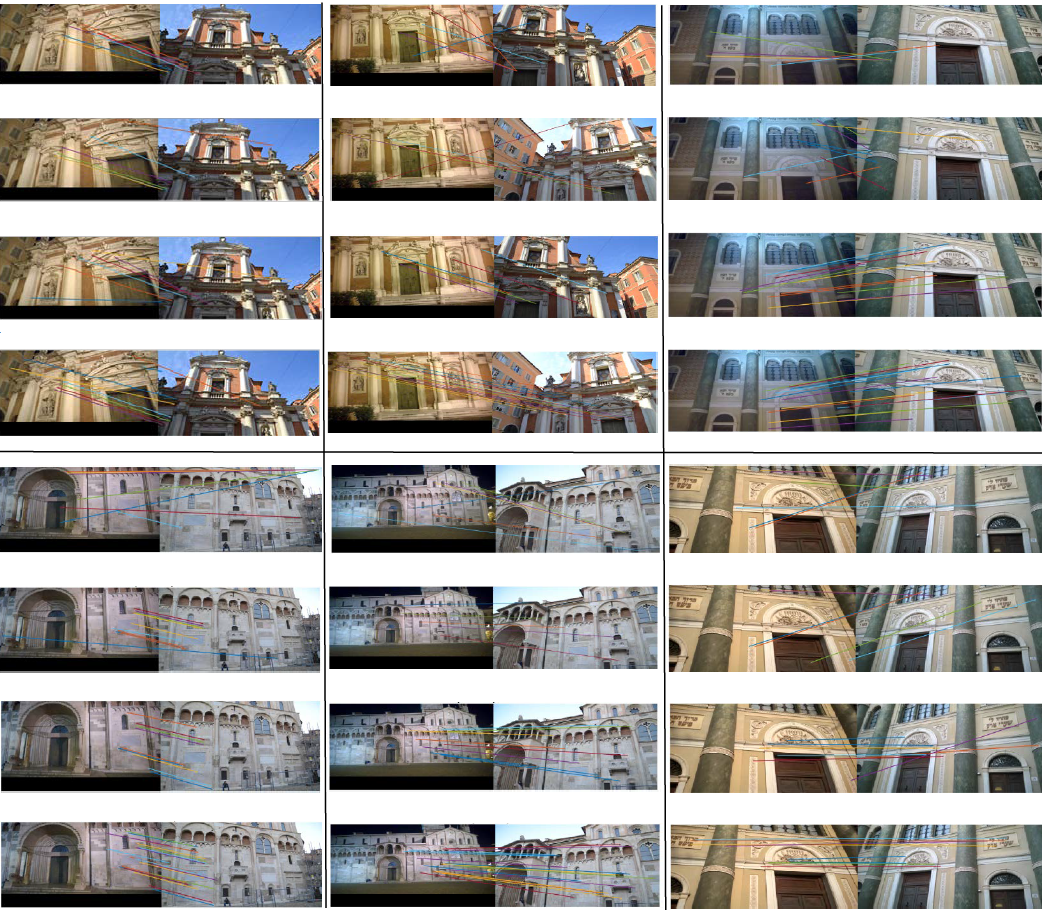}
	\caption{Samples of the matching results in scenarios where strong differences between query frame and reference image are present. Each cell of the image contains the results of the following methods, from up to botton: SIFT Baseline, Tokri \etal \cite{Torki2010}, Ours (SP), Ours Eq. \ref{eq_temporal_spatial}.}
	\label{fig:matches}
\end{figure*}

\section{Experimental results}

In order to experimentally validate the proposed method, we record a set of unconstrained first person videos covering four different historical buildings in the city of Modena: the Ducal Palace, the Cathedral, the St. George Church and the Synagogue. Furthermore, each scenario is composed of two sequences, one recorded during the day and one recorded during the night. To the best of our knowledge, this is the first collection of unconstrained egocentric videos of cultural heritage sites featuring both night and day sequences. To build the 3D models of the aforementioned buildings, we collect a set of images gathered from Flickr and use them in the reconstruction process. Since current methods employing 3D models for registration adopt daytime images only \cite{kroeger2014video, Irschara09fromstructure-from-motion-2009}, we also restrict our search to such images, maintaining the compatibility of our method to other approaches and providing a fair comparison. The final dataset is composed by a subset of the videos as queries, and a subset of the images used to build the 3D model as the references to match with. It consists of up to 50 query frames per building, obtained by selecting stable sequences using the pruning step described in Section \ref{sec:approach}; the 3D models are composed by 1150 images, 400k 3D points and 2000k descriptors. For each query frame, the previous $T$ frames are included to be used to enforce temporal consistency, with $T$ up to 20. Ground truth rotation and translation matrices for the query sequences have been obtained by manually aligning them to the 3D models, \ie by manually providing ground truth matches as input for a SfM tool \cite{wu2011visualsfm}. To allow further research on the egocentric video registration topic, we release the dataset featuring both query sequences and pre-built 3D models on the project page \footnote{http://imagelab.unimore.it/videoregistration.html}.

\subsection{Matching evaluation}

Matching images and video sequences acquired during the night with reference images featuring daytime lighting conditions is a challenging task, and we now show an evaluation of the performance of our method  under such circumstances. In the following, the query sequences are matched with a subset of the images used in the construction of the models and the number of correct matches is evaluated. This allows us to show the performance of different methods in terms of scored matches when dealing with steep changes in lighting conditions. Table \ref{tab:matches} compares a SIFT matching baseline (obtained using the VLFeat \footnote{http://www.vlfeat.org/} MATLAB libray), a recent matching method based on laplacian embedding of local features \cite{Torki2010} and three variations of our method. Note that in  \cite{Torki2010} the authors do not apply any spatial structure weighting. In particular, we evaluate the proposed approach in different steps of its pipeline in order to show the improvement that results from the refinement of the method as described in Sections \ref{sec:approach}. The first variation of our method evaluated reflects Eq. \ref{eq:singleframe}, where only SIFT and covariance descriptors are embedded into the space used to match images. The second and third variants of our method evaluated in Table \ref{tab:matches} are, respectively, the based on Eq. \ref{eq:singleframe} with the inclusion in the embedding space of the spacial similarity ($S_{i,j}^{T}$ in Eq. \ref{eq_temporal_spatial}) and the full method including the temporal consistency (Eq. \ref{eq_temporal_spatial}). In the table, the three variations of our method are referred to as: Ours Eq. \ref{eq:singleframe}, Ours (SP) and Ours Eq. \ref{eq_temporal_spatial}. These are the number of inlier matches scored for each query in each scenario (corresponding to the actual number of 2D-3D correspondences that will be fed to the PnP algorithm), and as the table shows the usage of our method can improve the baseline results significantly, especially when dealing with nighttime images. In particular, when matching daytime images, all the methods produce inlier ratios above 50\%; on the contrary, when challenged with queries acquired during the night and matching them with daytime reference images, the inlier ratios drastically decrease. In particular, while the SIFT baseline produces sufficient results during the day, a RANSAC loop on the matches computed during the night cannot formulate a viable transformation hypothesis (6\% inlier rate). Similarly, the method by Torki \etal and our initial approach based on Eq. 5 cannot cope with the noise in the descriptors that is due to the steep change in lighting conditions (respectively 20\% and 13\% inlier rates). On the other hand, adopting our robust spatial information and further extending it by embedding temporal robustness is shown to produce a matching where the inlier ratio is sufficiently high (62\%).

\begin{table}[ht]	
	\centering
	\resizebox{0.95\textwidth}{!}{
		\begin{tabular}{lccc|ccc}
			\hline
			& \multicolumn{3}{c|}{\textbf{Daytime}}                      & \multicolumn{3}{c}{\textbf{Nighttime}}                     \\
			\textbf{Method}                & \textbf{\# Inliers} & \textbf{\# Matches} & \textbf{Ratio} & \textbf{\# Inliers} & \textbf{\# Matches} & \textbf{Ratio} \\ \hline
			\textbf{Ours Eq. \ref{eq:singleframe}}     & 175                 & 334                 & 0.524          & 20                  & 155                 & 0.129          \\
			\textbf{Ours (SP)}               & 152                 & 216                 & 0.704          & 32                  & 58                  & 0.552          \\
			\textbf{Ours (SP+Temp)}          & 196                 & 262                 & 0.748          & 31                  & 50                  & 0.620          \\ \hline
			\textbf{SIFT Baseline}         & 178                 & 348                 & 0.511          & 12                  & 204                 & 0.059          \\
			\textbf{Torki \etal \cite{Torki2010}} & 234                 & 394                 & 0.594          & 28                  & 142                 & 0.197          \\ \hline
		\end{tabular}}
		\caption{Average number of inliers and matches per query.}
		\label{tab:matches}
	\end{table}

A parameter that has been considered during the experiments is the dimension of the resulting embedding space. Spectral Gap Analysis has been employed to determine the best embedding dimension for the employed dataset, but in order to evaluate the generality of the method we perform experiments evaluating the number of scored matches under varying embedding dimension. Table \ref{tab:embedding_dim} displays the result of this evaluation: it can be noticed how, while the value $60$ selected via spectral gap analysis provides the best performance, the variance is not significant. This result shows that the embedding dimension does not have a strong influence on the descriptor matching phase, hence the proposed method does not suffer from the presence of a strictly data-dependent parameter.

Another significant factor that impacts on the performance of the proposed approach is the size of the RoI from where to extract the covariance descriptor. In fact, sampling data from both an excessively small or broad  region would result in a descriptor which is, respectively, discriminative but less robust or less discriminative but more robust. Table \ref{tab:scalef} reports the results of our method (Eq. 7) under different patch sizes, where the base size (scale factor 1) of the patch is $[24,24] \times \psi_s(x_i)$ (scale of the corresponding SIFT keypoint). It can be noticed how lower values quickly degrade the number of inliers due to the loss in discriminative power of the covariance descriptor. Similarly, if the RoI is excessively broad (scale factors $> 2.5$) results in reduced due to significant overlap between the regions and a further loss in discriminative capabilities. In the following experiments, the size corresponding to a scale factor of 1 is adopted.

\begin{table}[ht]
	\centering
	
	\resizebox{0.7\textwidth}{!}{
		\begin{tabular}{lccc}
			\hline
			\textbf{Embedding Dimension} & \textbf{Ours Eq. \ref{eq:singleframe}} & \textbf{Ours (SP)} & \textbf{Ours Eq. 7} \\ \hline
			\textbf{20}                  & 126               & 134              & 181               \\ 
			\textbf{40}                  & 164               & 144              & 184               \\ 
			\textbf{60}                  & 175               & 152              & 196               \\ 
			\textbf{80}                  & 168               & 142              & 193               \\ 
			\textbf{100}                 & 177               & 145              & 179               \\ 
			\hline
		\end{tabular}}
		\caption{Average number of inliers scored by the variants of our method under different embedding dimension. }
		\label{tab:embedding_dim}
	\end{table}

	\begin{figure}[t]
		\centering
		\includegraphics[width=0.95\textwidth]{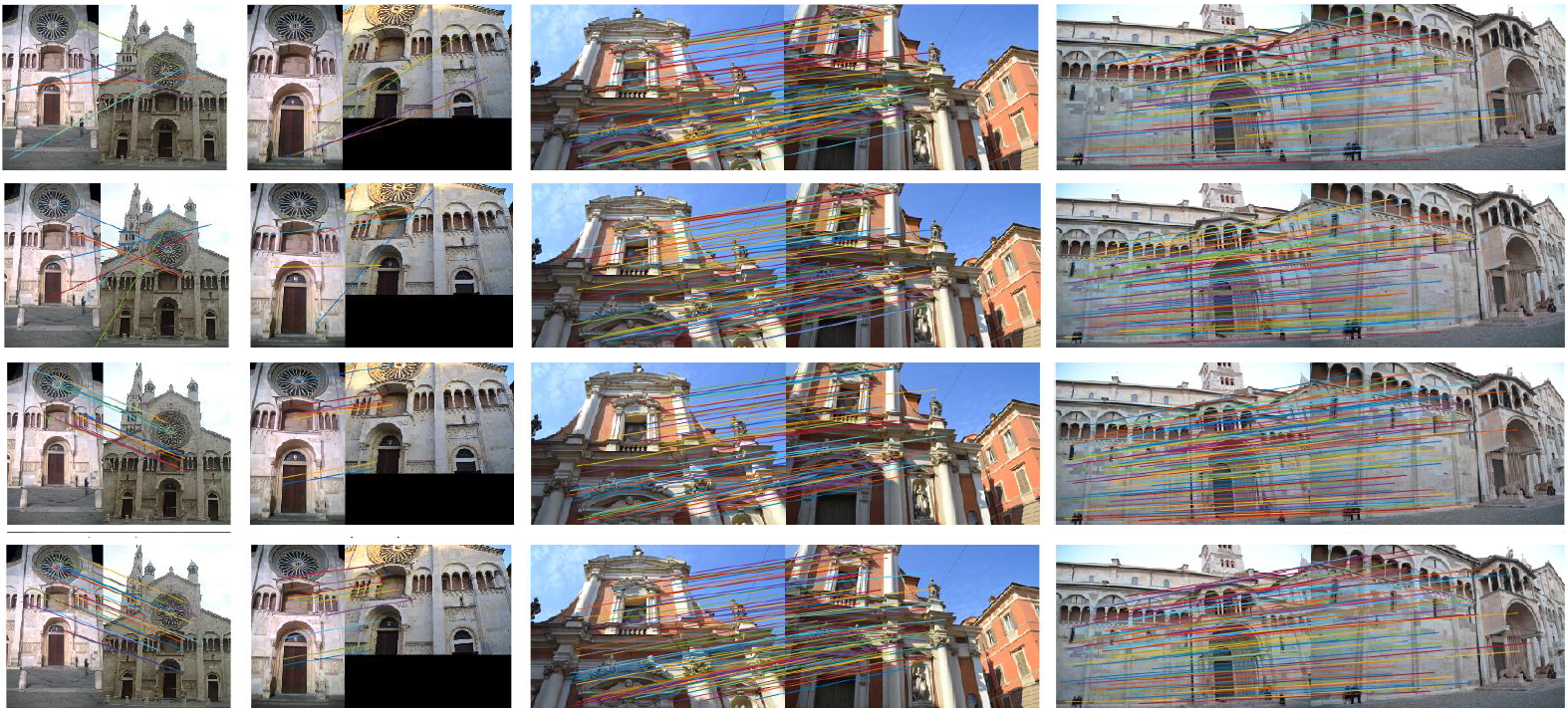}
		\caption{Samples of the matching results. From left to right: two matching frames extracted from night sequences compared to the model images and two matching frames extracted from day sequences. From up to bottom: SIFT Baseline, Tokri \etal \cite{Torki2010}, Ours (SP), Ours (SP+Temp).}
		\label{fig:matches}
	\end{figure}	
	
	\begin{table}[ht]
		\centering
		
		\resizebox{0.65\textwidth}{!}{
			\begin{tabular}{lccccccccc}
				\hline
				\textbf{Scale factor} & \textbf{0.7 } & \textbf{0.8 } & \textbf{0.9 }  & \textbf{1 } & \textbf{1.1 }  & \textbf{1.5 }  & \textbf{2.5 }  & \textbf{5.5 }  & \textbf{10 }  \\ \hline
				\textbf{\# Inliers}   & 4  & 49 & 195 & 196 & 196 & 180 & 175 & 157 & 118 \\ \hline
			\end{tabular}}
			\caption{Average number of inlier matches under different scales of the covariance RoI. }
			\label{tab:scalef}
		\end{table}

\subsection{Video Registration Analysis}

In the following experiments, we focus on the employment of our matching algorithm in the task of video registration, \ie the precise localization of the input egocentric video sequence on a 3D model. Given the wide range of possibilities implied by the adoption of the egocentric setting, videos acquired during both night and day are considered in the experiment. The goal is to obtain the extrinsic parameters of the camera acquiring the video at a frame level: similarly to \cite{kroeger2014video}, the quality of the performed registration is hence measured in terms of position RMS error (expressed in meters) and orientation RMS error (degrees). These two metrics express the quality of the registration, separately analyzing its two components, namely the obtained transition and rotation matrices. 

While a broader variety of methods have dealt with the task of registering single images, very few works consider the usage of video. Among them, the work by Kroeger and Van Gool \cite{kroeger2014video} represents the current state of the art in the task of registering videos to a 3D point cloud obtained via SfM. This method presents two significant differences respect to our setting: the cameras are fixed on a van instead of being head mounted, and it does not deal with the possibility of steep lighting changes. Despite this, a comparison against the work by Kroeger \etal is key in order to validate the performance of our method compared to what the current state of the art is. As a result, we perform an evaluation of the registration performance of both methods employing the same 3D models which only feature images acquired during the day, putting \cite{kroeger2014video} in its ideal working conditions. Since both approaches evaluate their results in terms of rotation and translation errors, the ground truth information of these matrices is also shared during the evaluation.

\begin{table}[t]
	\centering
	\resizebox{0.99\textwidth}{!}{\begin{minipage}{1.7\textwidth}\centering
			\begin{tabular}{ll|cc|cc|cc|cc|cc}
				\hline
				&                  & \multicolumn{2}{c|}{\textbf{Ours Eq. \ref{eq:singleframe}}} & \multicolumn{2}{c|}{\textbf{Ours (SP)}} & \multicolumn{2}{c|}{\textbf{Ours Eq. 7}} & \multicolumn{2}{c|}{\textbf{SIFT Baseline}} & \multicolumn{2}{c|}{\textbf{Kroeger \etal \cite{kroeger2014video}}} \\ \cline{3-12} 
				\multicolumn{2}{l|}{\textbf{Scenario}}                     & \textbf{Pos.}       & \textbf{Orient.}      & \textbf{Pos.}   & \textbf{Orient.}   & \textbf{Pos.}       & \textbf{Orient.}      & \textbf{Pos.}       & \textbf{Orient.}      & \textbf{Pos.}          & \textbf{Orient.}         \\ \hline
				\multirow{4}{*}{\textbf{Ducal Palace}} & \textbf{Day}     & 1.758               & 5.327                 & 0.809           & 2.572              & 0.648               & 2.518                 & 0.662               & 2.508                 & 0.596                  & 1.998                    \\
				& \textbf{\# Reg.} & \multicolumn{2}{c|}{19 / 20}                & \multicolumn{2}{c|}{19.2 / 20}        & \multicolumn{2}{c|}{19.5 / 20}               & \multicolumn{2}{c|}{19.8 / 20}              & \multicolumn{2}{c}{20 / 20}                      \\
				& \textbf{Night}   & 19.588              & 94.047                & 2.298           & 5.785              & 1.616               & 2.417                 & 17.350              & 59.600                & 3.912                  & 4.593                    \\
				& \textbf{\# Reg.} & \multicolumn{2}{c|}{20 / 25}                & \multicolumn{2}{c|}{23 / 25}          & \multicolumn{2}{c|}{24.8 / 25}               & \multicolumn{2}{c|}{11 / 25}                & \multicolumn{2}{c}{25 / 25}                      \\ \hline
				\multirow{4}{*}{\textbf{Cathedral}}    & \textbf{Day}     & 1.535               & 6.767                 & 0.7926          & 1.479              & 0.697               & 1.422                 & 0.761               & 1.600                 & 0.774                  & 6.6358                   \\
				& \textbf{\# Reg.} & \multicolumn{2}{c|}{24 / 25}                & \multicolumn{2}{c|}{24.8 / 25}        & \multicolumn{2}{c|}{24.8 / 25}               & \multicolumn{2}{c|}{25 / 25}                & \multicolumn{2}{c}{25 / 25}                      \\
				& \textbf{Night}   & 29.135              & 85.033                & 2.944           & 6.255              & 2.430               & 6.054                 & 22.407              & 64.042                & 4.850                  & 2.253                    \\
				& \textbf{\# Reg.} & \multicolumn{2}{c|}{21.6 / 25}              & \multicolumn{2}{c|}{23 / 25}          & \multicolumn{2}{c|}{24.6 / 25}               & \multicolumn{2}{c|}{24 / 25}                & \multicolumn{2}{c}{25 / 25}                      \\ \hline
				\multirow{4}{*}{\textbf{St. George}}   & \textbf{Day}     & 1.054               & 21.343                & 0.854           & 21.364             & 0.885               & 20.526                & 0.668               & 20.607                & 0.776                  & 16.686                   \\
				& \textbf{\# Reg.} & \multicolumn{2}{c|}{19.2 / 25}              & \multicolumn{2}{c|}{22.4 / 25}        & \multicolumn{2}{c|}{21 / 25}                 & \multicolumn{2}{c|}{23 / 25}                & \multicolumn{2}{c}{25 / 25}                      \\
				& \textbf{Night}   & 20.366              & 129.385               & 5.463           & 53.943             & 9.497               & 41.294                & 13.119              & 144.534               & 11.408                 & 86.663                   \\
				& \textbf{\# Reg.} & \multicolumn{2}{c|}{15.6 / 25}               & \multicolumn{2}{c|}{19 / 25}           & \multicolumn{2}{c|}{20.6 / 25}               & \multicolumn{2}{c|}{14.6 / 25}              & \multicolumn{2}{c}{25 / 25}                      \\ \hline
				\multirow{4}{*}{\textbf{Synagogue}}    & \textbf{Day}     & 9.621               & 68.855                & 5.860           & 41.723             & 0.976               & 24.626                & 2.432               & 32.467                & 0.841                  & 23.844                   \\
				& \textbf{\# Reg.} & \multicolumn{2}{c|}{19 / 25}                & \multicolumn{2}{c|}{14.4 / 25}        & \multicolumn{2}{c|}{15.7 / 25}               & \multicolumn{2}{c|}{24 / 25}                & \multicolumn{2}{c}{25 / 25}                      \\
				& \textbf{Night}   & 9.944               & 100.852               & 8.577           & 88.315             & 2.138               & 24.282                & 4.560               & 69.329                & 3.672                  & 30.717                   \\
				& \textbf{\# Reg.} & \multicolumn{2}{c|}{6 / 14}                 & \multicolumn{2}{c|}{7.2 / 14}         & \multicolumn{2}{c|}{12.6 / 14}                & \multicolumn{2}{c|}{9.8 / 14}              & \multicolumn{2}{c}{14 / 14}                      \\ \hline
			\end{tabular}
		\end{minipage}}
		\caption{Comparison of the video frame registration performance in different scenarios and lighting conditions. The position error is reported in meters, while the orientation error is reported in degrees.}
		\label{tab:registration}
	\end{table}

Table \ref{tab:registration} shows the results of this evaluation. In order to better study the correlation between improved matching and increased registration performance, in this experiment we also evaluate the variants of our method considered in Table \ref{tab:matches}, as long as the matching strategy proposed by \cite{Torki2010}. Except for \cite{kroeger2014video}, all the matching approaches evaluated are followed by the same ASPnP procedure, ensuring that changes in registration performance are due to the different inputs the algorithm receives, \ie matches of different quality. In order to provide comparable results with the method by Kroeger et al., results are expressed in terms of median error over 10 cross-validation iterations. In fact, since the smoothing component of \cite{kroeger2014video} always refines the initial PnP pose, it results in a root mean-squared (RMS) error one order of magnitude greater that the other methods in the presence of outlier poses (which tends to happen with nighttime sequences).

\begin{figure}[t]
	
	\centering
	\subfigure[]
	{\includegraphics[width=0.49\textwidth]{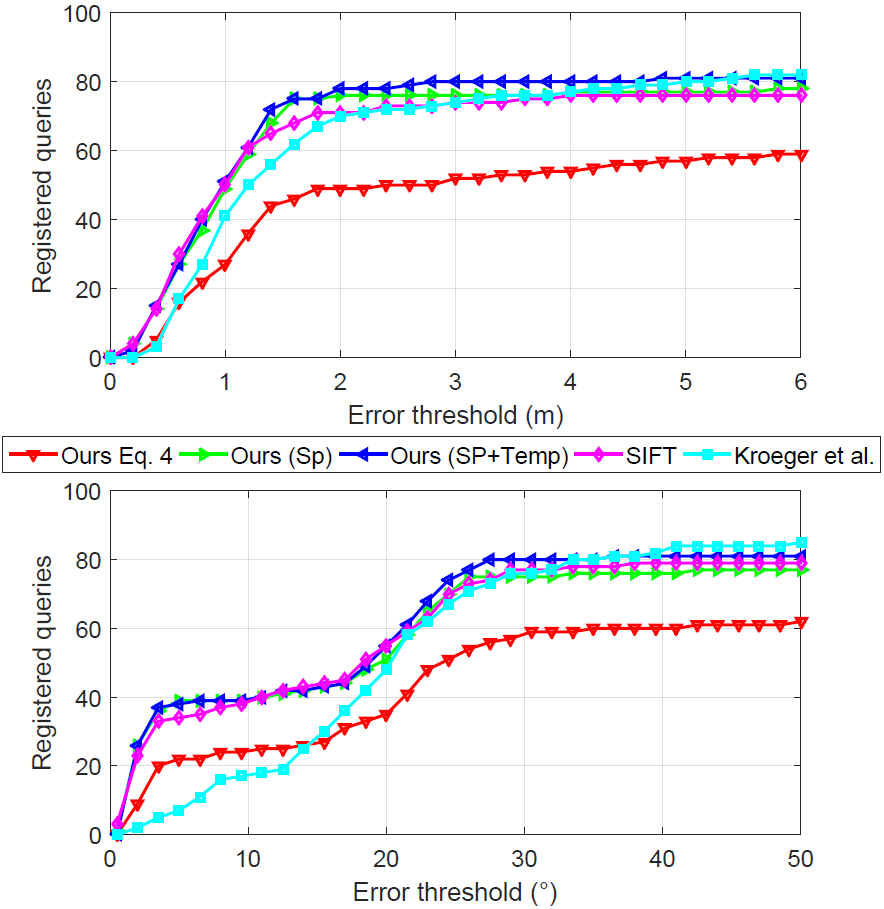}}
	\subfigure[]
	{\includegraphics[width=0.49\textwidth]{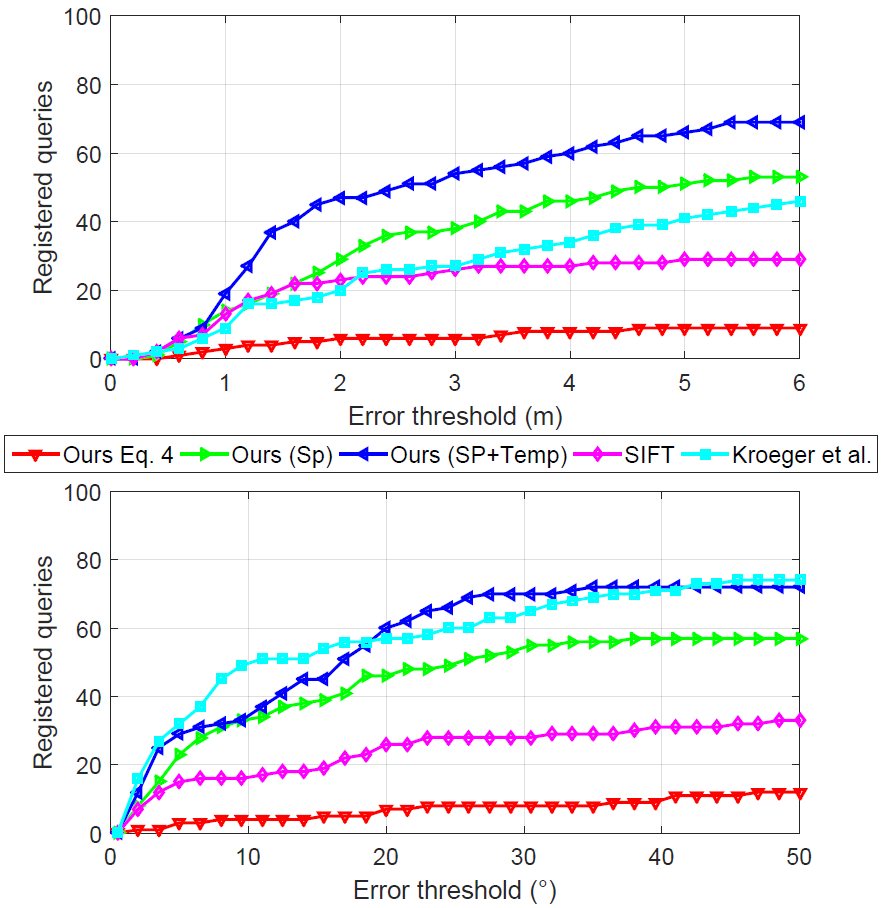}}

	\caption{Plots reporting the amount of registered query images under different rejection thresholds. On the left: daytime sequences. On the right: nighttime sequences. First row: position threshold (meters), second row: orientation threshold (degrees).}
	\label{fig:grafici}
\end{figure}

It can be noticed how, in general, the SIFT baseline cannot produce good registration results having a position error that is usually double or more the error of the other approaches and an orientation error that in average is around $90°$. This is due to the euclidean distance of SIFT descriptors not being able to produce sufficiently accurate matching results, preventing the RANSAC algorithm to correctly discriminate inliers and outliers, resulting in the failure of the ASPnP extrinsic parameters estimation. Analyzing the three variants of our method, it can be seen how employing both spatial and temporal consistency (Ours Eq. \ref{eq_temporal_spatial}) generally provides the best registration results despite having a slightly lower amount of matches. This confirms the fact that the temporal robustness of keypoints can remove actual outliers from the PnP procedure. As expected, the method by Kroeger \etal produces very good registration performance when dealing with video sequences acquired during the day but is less robust when facing significant differences in lighting conditions between input videos and 3D model. In particular, it suffers the most in terms of orientation error, with performance that can degrade from $16.686°$ during the day to $86.663°$ during the night in the scenario of the St. George. Finally, the matching strategy proposed by Torki \etal presents similar results, showing an increase in position and orientation errors between day and night that is due to the lower quality of matches produces. To better convey the results of the different methods on the individual sequences, Figure \ref{fig:grafici} reports the results of the registration phase in terms of number of registered images under different rejection thresholds. That is, the number of registered images obtained by rejecting all those registered with an error above a certain threshold. It can be noticed how, while in the left plot most of the methods quickly converge to the same registration performance, registering nighttime images (right plot) is significantly more difficult and the improvement achieved by our solution using both spatial and temporal information is significant.

As matching has become a crucial and time consuming part of 3D reconstruction, it is useful to analyze the different running times of the various methods adopted in the comparison. Table \ref{times} reports the results of this comparison, performed with un-optimized and un-parallelized MATLAB code on a i7-4790 CPU. It can be noticed how the SIFT baseline results in being the fastest method, albeit the least accurate one. On the other hand, more complex methods result in increased running times but better performance. Please notice that the method by Kroeger \etal performs a global optimization using all the queries at once, hence the increase in reported time. 

Considering the scalability of the proposed method, in order to register a query sequence to its 3D model the number of required 2D-3D correspondences is limited. This means that, provided a technique able to select a subset of the overall database of images, the execution time of the matching phase could remain constant despite the increase in scale of the 3D model, up to including thousands of images in the 3D reconstruction. In particular, we use the image retrieval approach presented in \cite{philbin2007object} to select the K images closer to the query (see Section 3) and perform the matching only using them. Performing the image retrieval preprocessing, i.e. codebook construction, description of the database images using the codebook and their ranking according to the similarity with the query requires the following time: 14.48 s for a database of 200 images, 23.54 s for 400 images, 56.14 s for 800, 166.00 s for 1600 and 398.11 s for 3200. Note that the reported times have been obtained using optimized C++ code (via the OpenCV library) and do not include feature (SIFT) extraction since it is a step that must be done when performing the 3D reconstruction regardless of the usage of a BoW technique. These results show that increasing the number of images in the database up to the thousands only has a small impact on the overall time requirements.

\begin{table}[ht]
	\centering
	
	\resizebox{1\textwidth}{!}{
	\begin{tabular}{ccccccc}
		\hline
		Method   & SIFT Baseline & Ours Eq. 4 & Ours (SP) & Ours Eq. 7 & Kroeger \etal \cite{kroeger2014video} & Torki \etal \cite{Torki2010} \\ \hline
		Time (s) & 78.42         & 98.28      & 102.27    & 105.48     & 288.11*                        & 100.16              \\ \hline
	\end{tabular}
	
}
\caption{Execution times (s) for registering a query sequence.}
\label{times}
\end{table}

\section{Conclusions}

In this paper we proposed a video frame registration approach that copes with the challenges of severe illumination changes that often occur in egocentric video sequences. The presented embedding function, that defines a feature space which encodes visual similarity, spatial arrangement of features and their stability over time, allows us to use standard techniques like bipartite graphs to robustly compute 2D-3D correspondence between the candidate frame and the pre-built 3D model. Experimental results demonstrate that the proposed approach obtains better performances with respect to the current state of the art in terms of video registration accuracy and show its robustness in unconstrained day/night video sequences. 

\section*{Acknowledgment}
This work was supported by the Fondazione Cassa di Risparmio di Modena project: ``Vision for Augmented Experiences'' (VAEX).





\section*{References}
\bibliographystyle{elsarticle-num} 
\bibliography{egbib}

\end{document}